\pdfoutput=1
\documentclass[nohyperref]{article}

\usepackage{microtype}
\usepackage{graphicx}
\usepackage{booktabs} 

\usepackage{hyperref}

\usepackage[accepted]{icml2023}

\usepackage{amsmath}
\usepackage{amssymb}
\usepackage{mathtools}
\usepackage{amsthm}
\usepackage{mathtools}
\usepackage{bm}
\usepackage{caption}
\usepackage{subcaption}
\usepackage{multirow}
\usepackage{siunitx}
\usepackage{threeparttable}

\usepackage[capitalize,noabbrev]{cleveref}
\usepackage{hyperref}

\theoremstyle{plain}

\theoremstyle{definition}

\theoremstyle{remark}

\DeclarePairedDelimiterX{\infdivx}[3]{(}{)}{%
  #1\;\delimsize\|\;#2;#3%
}
\newcommand{\infdiv}{\mathrm{KL}\infdivx}

\DeclareMathOperator*{\argmax}{arg\,max}

\usepackage[textsize=tiny]{todonotes}

\icmltitlerunning{Non-Uniform Neural Operator}

\begin{document}

\twocolumn[
\icmltitle{NUNO: A General Framework for \\ Learning Parametric PDEs with Non-Uniform Data}



\icmlsetsymbol{equal}{*}

\begin{icmlauthorlist}
\icmlauthor{Songming Liu}{thu}
\icmlauthor{Zhongkai Hao}{thu}
\icmlauthor{Chengyang Ying}{thu}
\icmlauthor{Hang Su}{thu,lab}
\icmlauthor{Ze Cheng}{comp}
\icmlauthor{Jun Zhu}{thu,lab}
\end{icmlauthorlist}

\icmlaffiliation{thu}{Dept. of Comp. Sci. and Tech., Institute for AI, THBI Lab, BNRist Center, Tsinghua-Bosch Joint ML Center, Tsinghua University}
\icmlaffiliation{comp}{Bosch Center for Artificial Intelligence}
\icmlaffiliation{lab}{Pazhou Laboratory (Huangpu), Guangzhou, China}

\icmlcorrespondingauthor{Jun Zhu}{dcszj@tsinghua.edu.cn}

\icmlkeywords{Machine Learning, ICML, Machine Learning for Physics}

\vskip 0.3in
]



\printAffiliationsAndNotice{}  

\begin{abstract}
The neural operator has emerged as a powerful tool in learning mappings between function spaces in PDEs. However, when faced with real-world physical data, which are often highly non-uniformly distributed, it is challenging to use mesh-based techniques such as the FFT. To address this, we introduce the Non-Uniform Neural Operator (NUNO), a comprehensive framework designed for efficient operator learning with non-uniform data. Leveraging a K-D tree-based domain decomposition, we transform non-uniform data into uniform grids while effectively controlling interpolation error, thereby paralleling the speed and accuracy of learning from non-uniform data. We conduct extensive experiments on 2D elasticity, (2+1)D channel flow, and a 3D multi-physics heatsink, which, to our knowledge, marks a novel exploration into 3D PDE problems with complex geometries. Our framework has reduced error rates by up to $60\%$ and enhanced training speeds by $2\times$ to $30\times$. The code is now available at \url{https://github.com/thu-ml/NUNO}.
\end{abstract}

\section{Introduction}
In many fields such as fluid dynamics \cite{cebeci2005computational} and electromagnetism \cite{de2014mathematical}, scientists and engineers often resort to numerical simulation rather than expensive experiments to study the behavior of physical systems under different parameters. However, traditional numerical methods can be too time-consuming for complex physical systems characterized by partial differential equations (PDEs) \cite{hao2022physics}. Neural operator, a class of data-driven surrogate models, approximates the mapping from the parameter function to the solution of PDEs. When inference, one forward pass is several orders of magnitude faster than traditional numerical methods \cite{li2020fourier}. 

\begin{figure}[!t]
\begin{center}
\centerline{\includegraphics[width=\columnwidth]{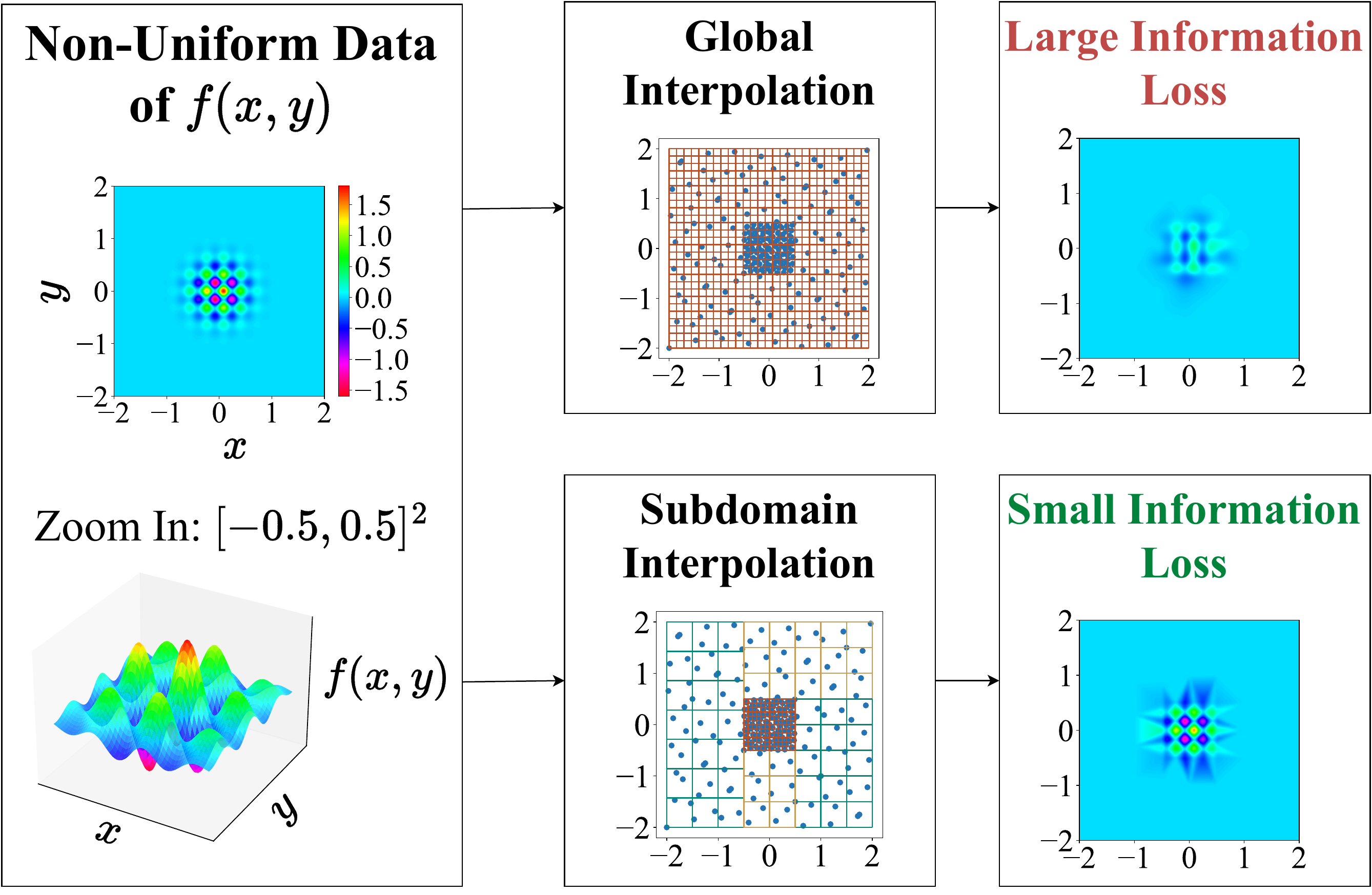}}
\caption{\textbf{Illustration of non-uniform data}. The physical quantity $f(x,y)$ varies greatly at the center but remains smooth in the far field, prompting a higher concentration of sensors (blue dots) in the center. Direct global interpolation may overlook these variations and yield a substantial error. Instead, our approach involves decomposing the domain into five subdomains and adaptively applying grids of varying sizes (denoted by different colored lines) to minimize error.}
\label{fig_intro}
\end{center}
\end{figure}

Among recently developed neural operators, mesh-based neural operators are typically fast and accurate, whose input and output are specified as uniform grids. For example, Fourier neural operator (FNO) \cite{li2020fourier} can efficiently extract highly nonlinear and nonlocal features and perform accurate predictions via the fast Fourier transform (FFT). Besides, a uniform grid is naturally compatible with a discretization of PDEs, which is the basis of many important works \cite{long2018pde, gin2021deepgreen, liu2022predicting}. However, real-world physical data are usually highly non-uniformly distributed (see Figure~\ref{fig_intro} for an illustration), and mesh-based methods are difficult to apply due to large interpolation errors. To overcome this limitation, some attempts to develop mesh-less neural operators \cite{lu2021learning,brandstetter2022message} at the expense of degradation in speed and loss of astounding properties such as the FFT and the equivalence between convolution and differentiation \cite{long2018pde}. Others have considered borrowing adaptive meshes from numerical methods to learn a bijection between point clouds (a general class of non-uniform data) and uniform grids \cite{li2022fourier}. But such an approach has not been theoretically justified yet and sometimes hand-crafted designs are needed. In a word, the challenge of non-uniform problems is: how to achieve a good trade-off between speed (to efficiently process unstructured data) and accuracy (to control interpolation error).

In this paper, we propose a non-uniform neural operator (NUNO) to facilitate mesh-based neural operators over complex geometries and highly non-uniform data. Specifically, we design a K-D tree algorithm for domain decomposition, and then adaptively use different-size grids on different subdomains for interpolation. The interpolated subdomain grids are then used to train the underlying mesh-based model. In this way, we convert unstructured data into uniform grids under the premise of the controlled interpolation error, so that many fast mesh-based techniques can be used. In addition, different-size subdomain grids induce the neural network to learn multi-scale features and improve its abstraction ability. 

In our experiments, we investigate three challenging problems of complex geometries, where non-uniform sensors are placed to capture complex physical phenomena: 2D elasticity from \citet{li2022fourier}, (2+1)D channel flow adapted from \citet{aparicio2020cfd}, and 3D heatsink from COMSOL application gallery \cite{comsol}, which is a fresh attempt to the 3D problems of complex geometries. The experimental results show that our method achieves state-of-the-art performance as well as fast speed, compared with comprehensive baselines including mesh-less and mesh-based neural operators (via global interpolation).

\paragraph{Contributions.} In summary, the general contributions of this paper include
\begin{itemize}
    \item We propose a general framework, a non-uniform neural operator (NUNO), for facilitating mesh-based operator learning with non-uniform data.
    \item We develop a K-D tree algorithm to iteratively divide the problem domain into several subdomains to reduce non-uniformity and subsequent interpolation error.
    \item Our method has achieved the best cost-accuracy trade-off. It lowers the error rates by $55\%$ in 2D elasticity, by $61\%$ in (2+1)D channel flow, and by $34\%$ in 3D heatsink. At the same time, it achieves promising speedups: at most $30\times$ speedups compared with mesh-free baselines and ${2\times}$ to ${4\times}$ speedups compared with mesh-based baselines. 
\end{itemize}

\section{Related Work}
In this section, we briefly discuss the related frontier work.

\paragraph{Traditional Numerical Methods.} Nowadays, numerical methods are still the most prevalent ones to solve PDEs, involving the finite element method (FEM) \cite{reddy2019introduction}, finite volume method (FVM) \cite{moukalled2016finite}, material point method (MPM), multi-grid methods \cite{hackbusch2013multi}, etc. They suffer from the ``curse of dimensionality" \cite{xue2020amortized}, despite their high accuracy and complete theoretical foundation. A moderately complex physical system can cost several minutes to simulate, rendering their infeasibility in scenarios like optimization where PDEs are solved frequently.


\paragraph{Mesh-based vs. Mesh-less Neural Operators.}
Mesh-based neural operators are a class of fast, flexible, and accurate neural operators in which both the input (parameter function) and output (solution to the PDEs) are discretized into uniform grids. Famous examples include convolutional neural operators \cite{gao2021phygeonet,khara2021field}, Fourier neural operator (FNO) \cite{li2020fourier}, and Transformer neural operators \cite{cao2021choose,hao2023gnot}. In contrast, mesh-less neural operators such as DeepONet \cite{lu2021learning} and graph neural operators \cite{sanchez2020learning,li2020neural} allow for arbitrary inputs and queries, which can be used in non-uniform data. But they have lost the speed and efficient mesh-based techniques like the FFT. Besides, other lines of neural network-based PDE solvers involve physics-informed neural networks (PINNs) \cite{raissi2019physics} and neural fields \cite{sitzmann2020implicit}.

\paragraph{Deformable Meshes.}
Adaptive mesh methods \cite{babuvska1990p,huang2010adaptive} come from the community of numerical methods, which can deform between irregular meshes and uniform meshes. Some have attempted to learn such a deformation to facilitate mesh-base neural operators \cite{li2022fourier}. However, in addition to the lack of theoretical guarantee, it is extremely difficult for a neural network to learn such a mapping, as there is no corresponding loss function to evaluate the progress and guide its learning towards a uniform grid.

\section{Methodology}
In this section, we will first give a formulation of the problem considered in this paper. We then present a K-D tree algorithm for domain decomposition, followed by an overall pipeline of our method, a non-uniform neural operator.


\subsection{Problem Formulation}
We consider a system of PDEs parameterized by some function $a(x)\in \mathcal{A}, x\in \Omega_a$. When the parameter is specified to be $a(x)$, the solution to the PDEs is denoted as $u(y) \in\mathcal{U}, y\in\Omega_u$. Here, $\mathcal{A}$ and $\mathcal{U}$ are two Banach function spaces with domains of definition $\Omega_a\subset\mathbb{R}^{d_a}$ and $\Omega_u\subset\mathbb{R}^{d_u}$, respectively, where $d_a$ and $d_u$ are two positive integers. Given a dataset $\{ (a_j, u_j) \}_{j=1}^N$ where $a$ follows some distribution $\mu$ and $a,u$ are both represented as point clouds $a:=\{ a(x^{(i)}) \}_{i=1}^M, u:=\{ u(y^{(i)}) \}_{i=1}^M$\footnote{Here, we use the same notation for the function $a,u$ and their point-cloud values, which might be seen as a notation abuse.}, we hope to approximate the ground-truth operator $G^\dagger\colon \mathcal{A} \rightarrow \mathcal{U}$ or equivalently, $G^\dagger\colon a(x) \mapsto u(y)$ with a \emph{mesh-based} neural operator $G_{\theta}, \theta\in\Theta$ for some parameter space $\Theta$.

To capitalize on the speed and flexibility of mesh-based methods like the FFT, we interpolate the unstructured point clouds $a$ and $u$ into uniform grids, yielding vectors $\mathbf{a}$ and $\mathbf{u}$ respectively. Let $\phi$ denote a linear reconstruction or backward interpolation from the uniform grids to the point cloud. The prediction error can then be approximated using the following equation:
\begin{equation}
\begin{aligned}
    \| \phi(G_{\theta}(\mathbf{a})) - u \|
    \leq& \| \phi(G_{\theta}(\mathbf{a}) - \mathbf{u}) \| 
    + \| \phi(\mathbf{u}) - u \|\\
    \leq& \|\phi\| \underbrace{\| G_{\theta}(\mathbf{a}) - \mathbf{u} \|}_{\text{prediction error}} 
    + \underbrace{\| \phi(\mathbf{u}) - u \|}_{\text{interpolation error}},
\end{aligned}
\end{equation}
where $\|\cdot\|$ is the $L^2$ norm. The first term represents the prediction error of the neural operator, contingent upon the model's capability and the optimization's progression. Furthermore, the interpolation from $a$ to $\mathbf{a}$, can also negatively influence this term. The information lost during the interpolation process can potentially degrade the model's performance. The second term signifies the error of the interpolation from $\mathbf{u}$ back to $u$, which can directly impact the overall accuracy. Upon this analysis, it is evident that interpolation errors hold significant sway over the total error.

\paragraph{Domain Decomposition.}
In cases of highly non-uniform data (refer to Figure~\ref{fig_intro} for an example), the interpolation error can at times overshadow the total error. In our endeavor to mitigate this particular error, we opt to decompose the domains $\Omega_a$ and $\Omega_u$ into several subdomains, expressed as $\Omega_a=\bigcup_{k=1}^{n_a} \Omega_a^{(k)}, \Omega_u=\bigcup_{k=1}^{n_u} \Omega_u^{(k)}$. Within these subdomains, we perform interpolation using varied grid sizes to procure uniform grids, denoted as $\{ \mathbf{a}^{(k)} \}_{k=1}^{n_a}, \{ \mathbf{u}^{(k)} \}_{k=1}^{n_u}$. By judiciously decomposing the domain into subdomains, we can adaptively apply denser meshes in areas characterized by a high point density, and conversely, coarser meshes in areas where the points are more sparsely distributed.

\paragraph{Optimization Target.}
Formally speaking, with the domain decomposition strategy, our optimization target can be described as:
\begin{equation}
    \min_{\theta\in\Theta} \mathbb{E}_{a\sim \mu}  
    \left\| \phi \left( G_{\theta}(\{ \mathbf{a}^{(k)} \}_{k=1}^{n_a}) \right) - u \right\|,
\end{equation}
where $\{ \mathbf{a}^{(k)} \}_{k=1}^{n_a}$ is interpolated from $a$ and $G_{\theta}$ takes $\{ \mathbf{a}^{(k)} \}_{k=1}^{n_a}$ as input and outputs a prediction for $\{ \mathbf{u}^{(k)} \}_{k=1}^{n_u}$. Finally, the output of $G_{\theta}$ is interpolated back to the point cloud to obtain a prediction for $u$.

\begin{algorithm}[tb]
   \caption{Domain Decomposition via a K-D Tree}
   \label{alg_1}
\begin{algorithmic}[1]
   \STATE {\bfseries Input:} initial point cloud $D^{(0)}$ and the number of sub-point clouds $n$
   \STATE {\bfseries Output:} a sef of sub-point clouds $\mathcal{S}$
   \STATE {\bfseries Initialize:} $\mathcal{S} \leftarrow \{ D^{(0)} \}$
   \REPEAT
   \STATE Choose $D^* = \argmax_{D\in \mathcal{S}} (|D| \cdot \infdiv{P}{Q}{D})$
   \STATE $\mathcal{S}\leftarrow\mathcal{S}-\{ D^* \}$
   \STATE Select the dimension $k, 1\le k \le d$, where the bounding box of $D^*$ has the largest scale
   \STATE Determine the hyperplane $x_k = b^*$, where $b^* = \argmax_{b\in \mathcal{B}} \mathrm{Gain}(D^*, b)$ and $\mathcal{B}$ is a set of discrete candidates for $b$
   \STATE Partition $D^*$ with $x_k = b^*$
   \STATE $\mathcal{S}\leftarrow\mathcal{S}\cup\{ D^*_{x_k>b^*}, D^*_{x_k\le b^*} \}$
   \UNTIL{$|\mathcal{S}|=n$}
\end{algorithmic}
\end{algorithm}

\begin{figure*}[htbp]
\begin{center}
\centerline{\includegraphics[width=\textwidth]{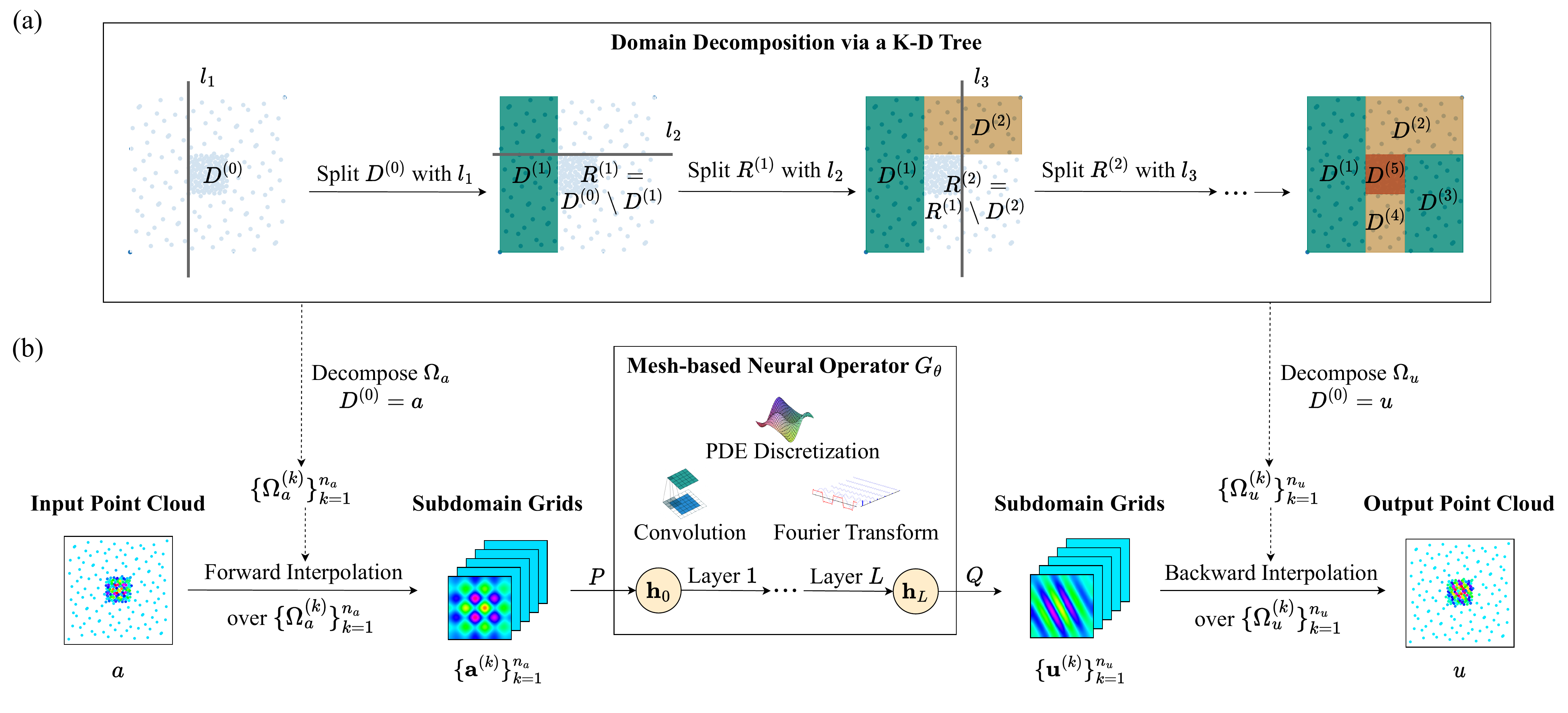}}
\caption{\textbf{(a) Domain decomposition:} starting from $D^{(0)}$, each time we choose to split a sub-point cloud with a hyperplane $l_i$. After $4$ iterations, we obtain $5$ sub-point clouds distributed more uniformly within their bounding boxes. \textbf{(b) NUNO framework:} 1. interpolation to subdomain grids; 2. projection by $P$; 3. pass through the mesh-based neural operator with $L$ layers ($\mathbf{h}_0,\dots,\mathbf{h}_L$ indicate hidden embeddings of each layer); 4. projection by $Q$; 5. interpolation back to the point cloud.}
\label{fig:nuno}
\end{center}
\vskip -0.2in
\end{figure*}

\subsection{Domain Decomposition via a K-D Tree}\label{sec_kd}
Domain decomposition aims to approximate the distribution of points within a given point cloud by uniform distributions defined in a series of subdomains, which corresponds to varied uniform subdomain grids. Upon obtaining this approximation, we separately perform interpolation in each subdomain, using a grid size proportional to the number of points living in it to maintain a low interpolation error. 

Let $D^{(0)}=\{ x^{(i)}:= (x_1^{(i)}, \dots, x_d^{(i)}) \}_{i=1}^m$ be an initial point cloud, where $d$ and $m$ represents the dimensionality and the number of points in the point cloud, respectively. The optimization target of domain decomposition can be framed as minimizing the approximation error, which is quantified by the Kullback-Leibler (KL) divergence:
\begin{equation}\label{eq:kdtarget}
    \min_{D^{(1)}, \dots, D^{(n)}} \sum_{i=1}^n \frac{|D^{(i)}|}{|D^{(0)}|}  \infdiv{P}{Q}{D^{(i)}},
\end{equation}
where $\bigcup_{i=1}^{n} D^{(i)}=D^{(0)}$ constitutes a $n$-partition of $D^{(0)}$ with disjoint bounding boxes. A bounding box is defined as the smallest enclosure containing all the points. For instance, the bounding box for $D^{(0)}$ is given by $[x_{\mathrm{min},1},x_{\mathrm{max},1}]\times\cdots \times [x_{\mathrm{min},d},x_{\mathrm{max},d}]$, where $x_{\mathrm{min},j} = \min \{ x_j^{(i)} \}_{i=1}^m$ and $x_{\mathrm{max},j} = \max \{ x_j^{(i)} \}_{i=1}^m$. Additionally, $\infdiv{P}{Q}{D}$ is the KL divergence between the distribution of points in a point cloud $D$, denoted by $P$, and the uniform distribution within the bounding box of $D$, denoted by $Q$: 
\begin{equation}\label{eq_uniform}
    \infdiv{P}{Q}{D} := \sum_{j} \frac{|D_{j}|}{|D|} \ln\left(\frac{|D_{j}|}{|D|} \bigg/ \frac{1}{\prod_{i=1}^d N_i}\right).
\end{equation}
In calculating $P$, we employ histogram density estimation. Specifically, we divide the bounding box of $D$ uniformly into $N_1\times\cdots\times N_d$ cells, and $D_{j}$ is defined as $\{ x\mid x\in D \land x \in \text{the $j$-th cell} \}$, for $1\le j\le \prod_{i=1}^d N_i$. 


While the optimization problem in Eq.~\eqref{eq:kdtarget} is reduced to an integer programming issue and is NP-hard, we have devised a heuristic method to approximate the solution. This method is inspired by the K-D tree algorithm, recognized for its quasilinear complexity and compatibility with arbitrary point clouds. In each iteration, we select a sub-point cloud $D^*$ with the highest total KL divergence and split it using a hyperplane $x_k=b^*$. The parameter $k$ is selected such that the bounding box of $D^*$ has the largest scale. The chosen hyperplane maximizes the gain of the KL divergence, represented by $\mathrm{Gain}(D^*, b):=$
\begin{equation}\label{eq_uniform_after}
\begin{aligned}
    \infdiv{P}{Q}{D^*} &- 
    \frac{\left|D^*_{x_k>b}\right|}{|D^*|} \infdiv{P}{Q}{D^*_{x_k>b}} \\
    &- \frac{\left|D^*_{x_k\le b}\right|}{|D^*|} \infdiv{P}{Q}{D^*_{x_k\le b}},
\end{aligned}
\end{equation}
where $D^*_{x_k>b},D^*_{x_k\le b}$ are defined as $\{ x \mid x\in D^* \land x_k > b \}$ and $\{ x \mid x\in D^* \land x_k \le b \}$, respectively. Through the recursive application of this process, we can partition $D^{(0)}$ into $n$ sub-point clouds, decreasing the approximation error. We provide an outline of the algorithm in Algorithm~\ref{alg_1} and showcase it in Figure~\hyperref[fig:nuno]{2a}.

\paragraph{Time and Space Complexity.}
Under proper assumptions, the time and space complexity are, respectively, derived as $\mathcal{O}(nm\log m + nd)$ and $\mathcal{O}(md)$, where $n$ is the number of sub-point clouds, $m$ is the number of points in the initial point cloud, and $d$ is the dimensionality. We leave a detailed analysis in Appendix~\ref{app_complexity}. In practice, the value of $n$ acts as a cost-accuracy trade-off. As discussed in Section~\ref{sec_exp}, we have empirically found that $n\le 10^2$ are sufficient for general complex systems. In this way, the algorithm takes only a few seconds, which is completely negligible compared to the training time of neural operators.

\subsection{Non-Uniform Neural Operator}\label{sec_nuno}
In this subsection, we elucidate our framework for the non-uniform neural operator, designed to synergize rapid computation with high precision. As illustrated in Figure~\hyperref[fig:nuno]{2b}, the framework can be methodically deconstructed into five primary steps:

\begin{enumerate}
\item Initially, the input point-cloud values, denoted as $a$, are interpolated into input subdomain grids which are subsequently harmonized to a uniform size. The resulting dimension from this step is denoted as $n_a\times s$, where $s$ represents the padded mesh size.
\item The subdomain grids are subsequently transformed into hidden embeddings of dimension $n_h\times s$ via a mapping function $P\colon \mathbb{R}^{n_a}\rightarrow \mathbb{R}^{n_h}$. Here, $n_h$ symbolizes the hidden width. This mapping function is parameterized by a shallow fully connected layer, with parameters that are shared across all mesh locations.
\item These generated hidden embeddings are then propagated through the hidden layers of the underlying mesh-based neural operator.
\item Post this operation, the hidden embeddings are transformed into the output subdomain grids of size $n_u\times s$, facilitated by another mapping function $Q\colon \mathbb{R}^{n_h}\rightarrow \mathbb{R}^{n_u}$. This function is also parameterized by a shallow fully connected layer, with parameters shared across all mesh locations.
\item Lastly, these output subdomain grids are interpolated back into the output point cloud $u$, which constitutes the final prediction.
\end{enumerate}

\paragraph{Outstanding Cost-Accuracy Balance.}
By transforming the point cloud into structured data through domain decomposition, we not only maintain a handle on the interpolation error but also expedite operations such as convolution and the FFT, bolstering our ability to extract non-local and non-linear features. Additionally, for the same total size, the subdomain grid is of smaller size compared to the global grid. As a result, the size of hidden embeddings (i.e., $s$) diminishes, yielding a more compact representation and thus a significantly lower space-time cost.

\begin{figure*}[!t]
     \centering
     \begin{subfigure}[b]{0.33\textwidth}
         \centering
         \includegraphics[height=0.7\textwidth]{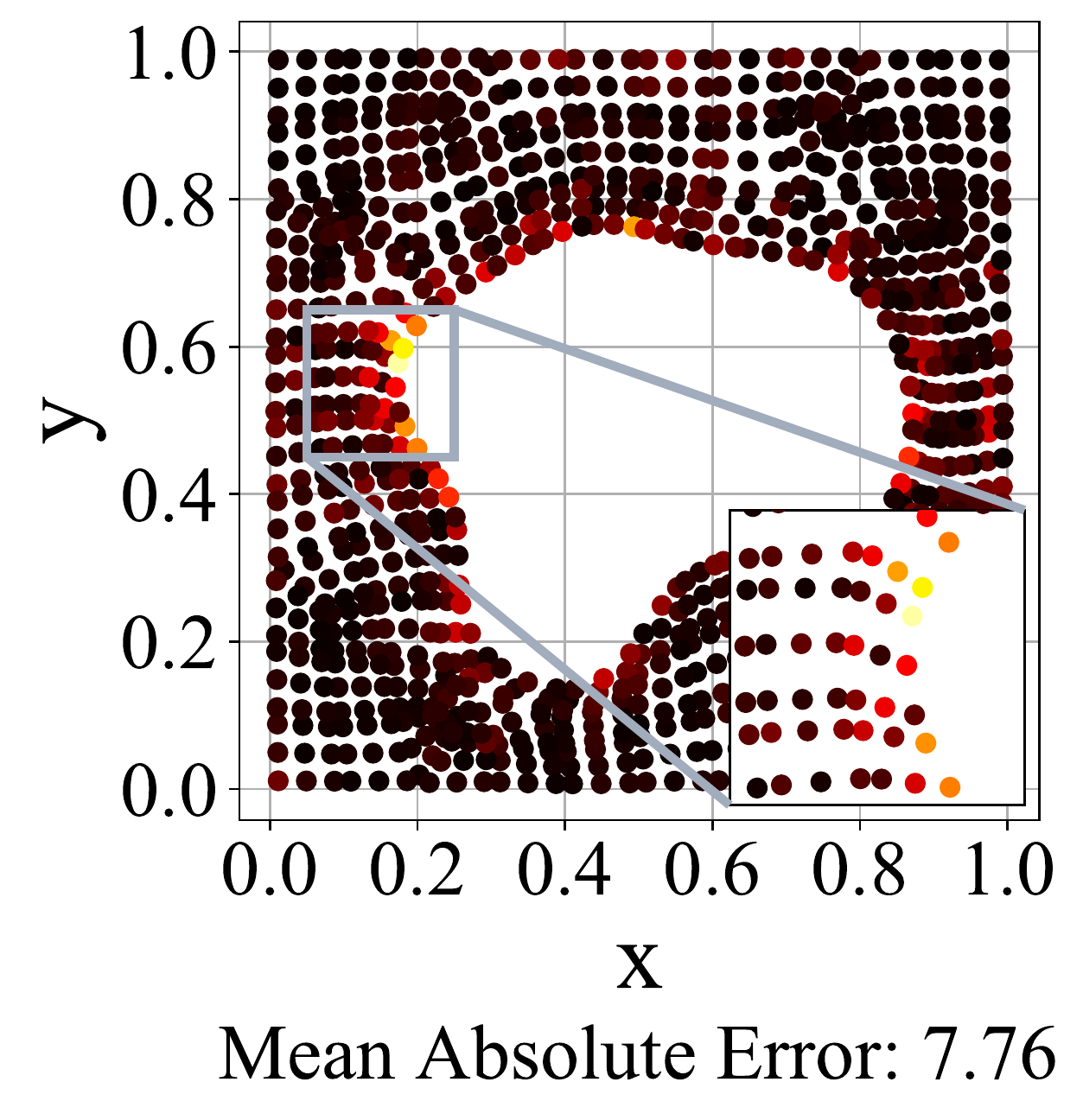}
         \caption{FNO (global interpolation)}
         \label{fig_exp1_fno}
     \end{subfigure}
     \hfill
     \begin{subfigure}[b]{0.33\textwidth}
         \centering
         \includegraphics[height=0.7\textwidth]{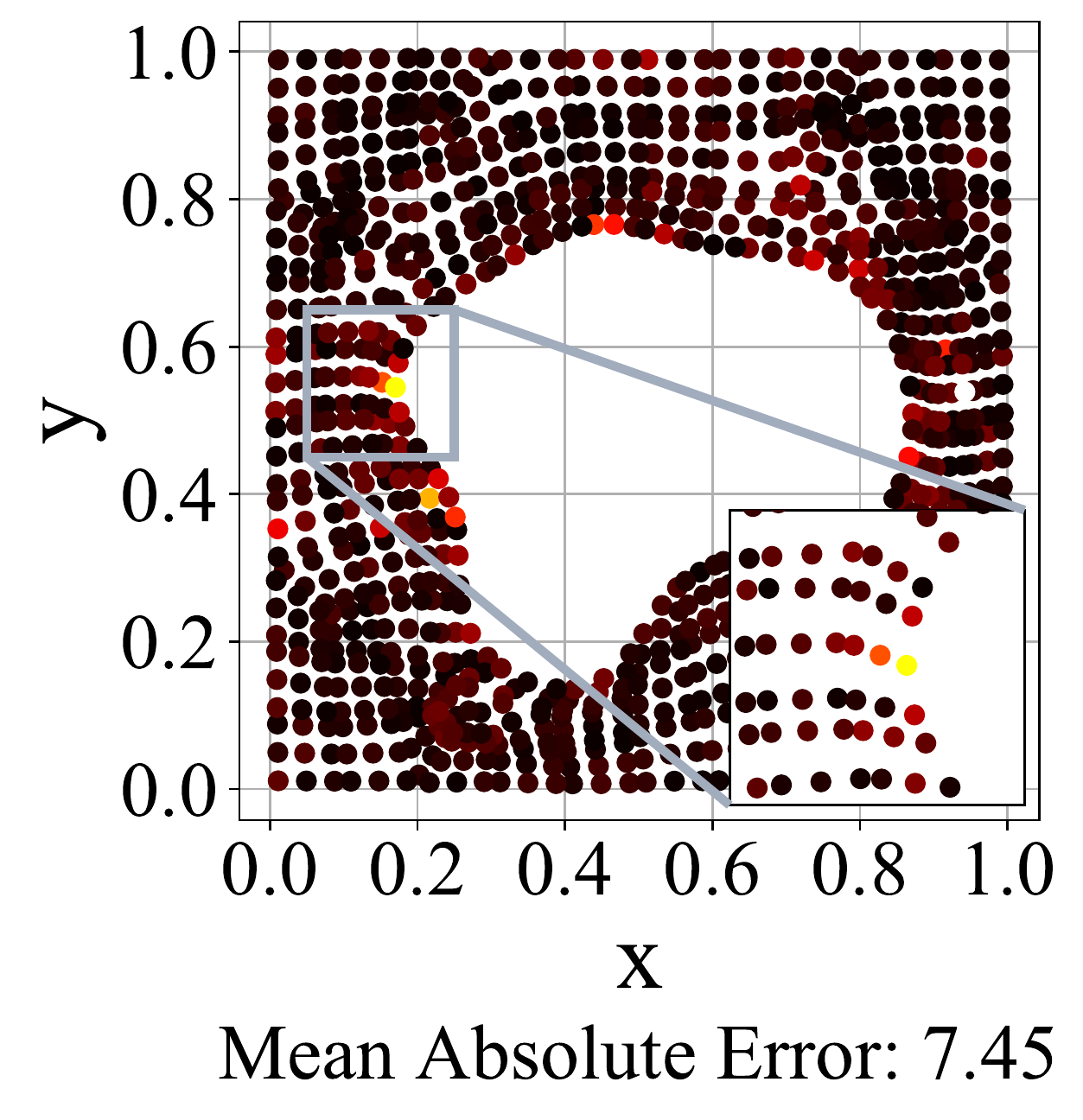}
         \caption{Geo-FNO (learned)}
         \label{fig_exp1_geofno}
     \end{subfigure}
     \hfill
     \begin{subfigure}[b]{0.33\textwidth}
         \centering
         \includegraphics[height=0.7\textwidth]{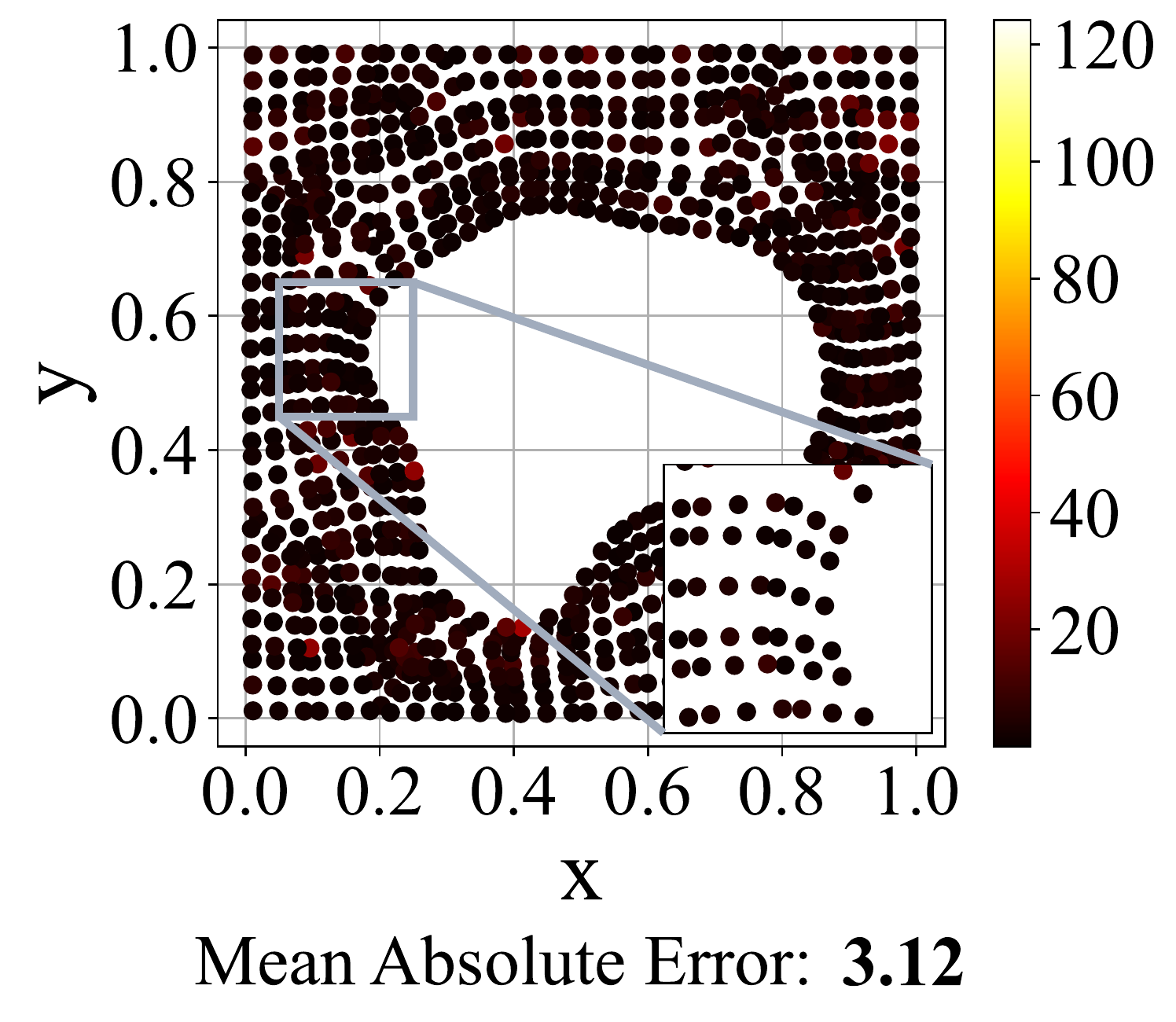}
         \caption{NU-FNO (\textbf{ours})}
         \label{fig_exp1_nufno}
     \end{subfigure}
\caption{Visualization of 2D elasticity: the absolute prediction error of $\bm{\sigma}$ at point-cloud locations.}
\label{fig_exp1}
\end{figure*}

\begin{table*}[!t]
\centering
\caption{Experimental results of 2D elasticity. R mesh and O mesh are adaptive meshes \cite{li2022fourier}.}
\renewcommand{\arraystretch}{1.2}
\begin{small}
\begin{threeparttable}
\begin{tabular}{l|l|l|l|l|l}
\hline
\multirow{2}{*}{\textbf{Method}} &
  \multirow{2}{*}{\textbf{Mesh Size}} &
  \multicolumn{2}{l|}{\textbf{Training Time}} &
  \multicolumn{2}{l}{\textbf{$L^2$ Relative Error ($\times 10^{-2}$)}} \\ \cline{3-6} 
 &
   &
  \multicolumn{1}{l|}{\textbf{per Epoch}} &
  \textbf{per Run} &
  \multicolumn{1}{l|}{\textbf{Training}} &
  \textbf{Testing} \\ \hline
NU-FNO (\textbf{ours})         & $1024$\tnote{1}     & \qty{2.3}{\s}  & \qty{19.6}{\min}  & $\bm{1.68} \pm \bm{0.08}$ & $\bm{1.93} \pm \bm{0.11}$ \\
FNO (global interpolation)   & $1681$     & $\bm{1.2} \,\mathrm{s}$  &  $\bm{9.7} \,\mathrm{min}$   & $3.41 \pm 0.08$          & $6.16 \pm 0.16$          \\
U-Net (global interpolation) & $1681$     & \qty{2.3}{\s}  &  \qty{18.8}{\min}  & $1.74 \pm 0.02$          & $6.82 \pm 0.06$          \\
Geo-FNO (R mesh)      & $1681$     & \qty{1.7}{\s}  & \qty{14.2}{\min}   & $3.53 \pm 0.09$          & $5.03 \pm 0.09$          \\
Geo-FNO (O mesh)      & $1353$     & \qty{2.0}{\s}  & \qty{16.4}{\min}   & $4.12 \pm 0.16$          & $4.28 \pm 0.15$          \\ \hline
Geo-FNO (learned)     & $\mathrm{meshless}$ & \qty{2.3}{\s}  &  \qty{19.1}{\min}  & $2.07 \pm 0.99$          & $4.31 \pm 2.24$          \\
GraphNO               & $\mathrm{meshless}$ & \qty{96.8}{\s} &  \qty{5.4}{\hour} & $17.5 \pm 1.26$         & $16.9 \pm 1.28$         \\
DeepONet              & $\mathrm{meshless}$ & \qty{40.0}{\s} &  \qty{11.0}{\hour} & $2.80 \pm 0.13$          & $12.0 \pm 0.16$         \\ \hline
\end{tabular}
\begin{tablenotes}
\item[1] In this paper, all mesh sizes mentioned for our method specifically refer to the \emph{combined total number} of points across all subdomains, rather than the number of points within each individual subdomain.
\end{tablenotes}
\end{threeparttable}
\end{small}
\label{tab_exp1}
\end{table*}

\paragraph{Versatility.}
Our framework exhibits compatibility with all mesh-based neural operators, necessitating no additional modifications for implementation. Moreover, domain decomposition facilitates flexibility in usage. For instance, we can execute different normalizations across varied subdomains, an approach particularly beneficial for multi-scale and discontinuous problems \cite{pang2020npinns}.

\paragraph{Alignment.} We note that the subdomain grids $\{ \mathbf{a}^{(k)} \}_{k=1}^{n_a}$ may differ in size, presenting challenges for batching and propagation. This issue is circumvented by aligning the grids to a uniform size. This alignment can be achieved by employing size-insensitive structures like the SPP-Net \cite{he2015spatial} or the Transformer \cite{vaswani2017attention} to generate embeddings of a consistent size. Alternatively, we could directly resize all the subdomain grids to match in size using up/down sampling or FFT-IFFT methods, a common technique in computer vision that we utilize in our implementation. Further details are elaborated in Appendix~\ref{app_exp}.

\begin{figure*}[h]
     \centering
     \begin{subfigure}[b]{0.49\textwidth}
         \centering
         \includegraphics[height=0.3\textwidth]{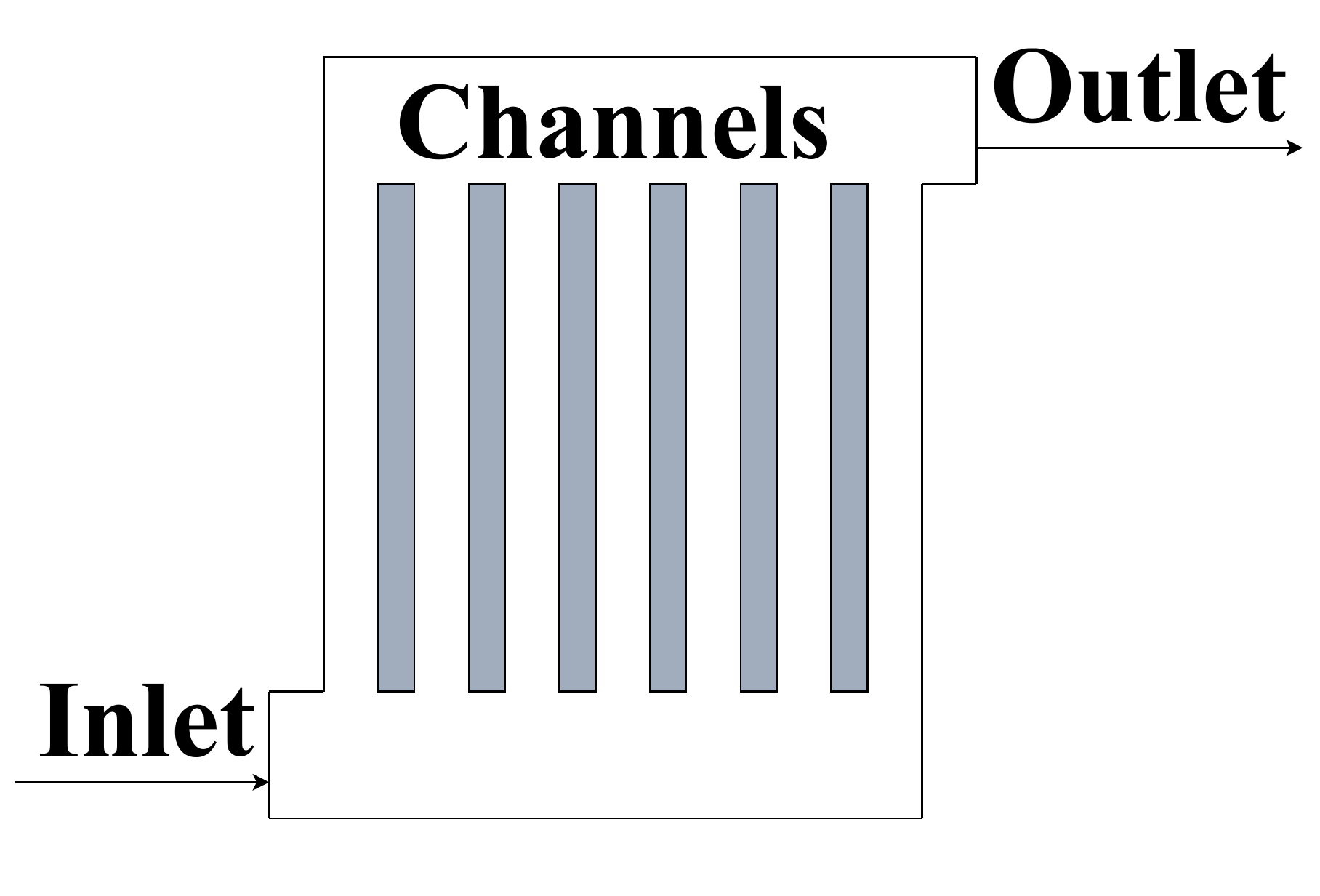}
         \includegraphics[height=0.3\textwidth]{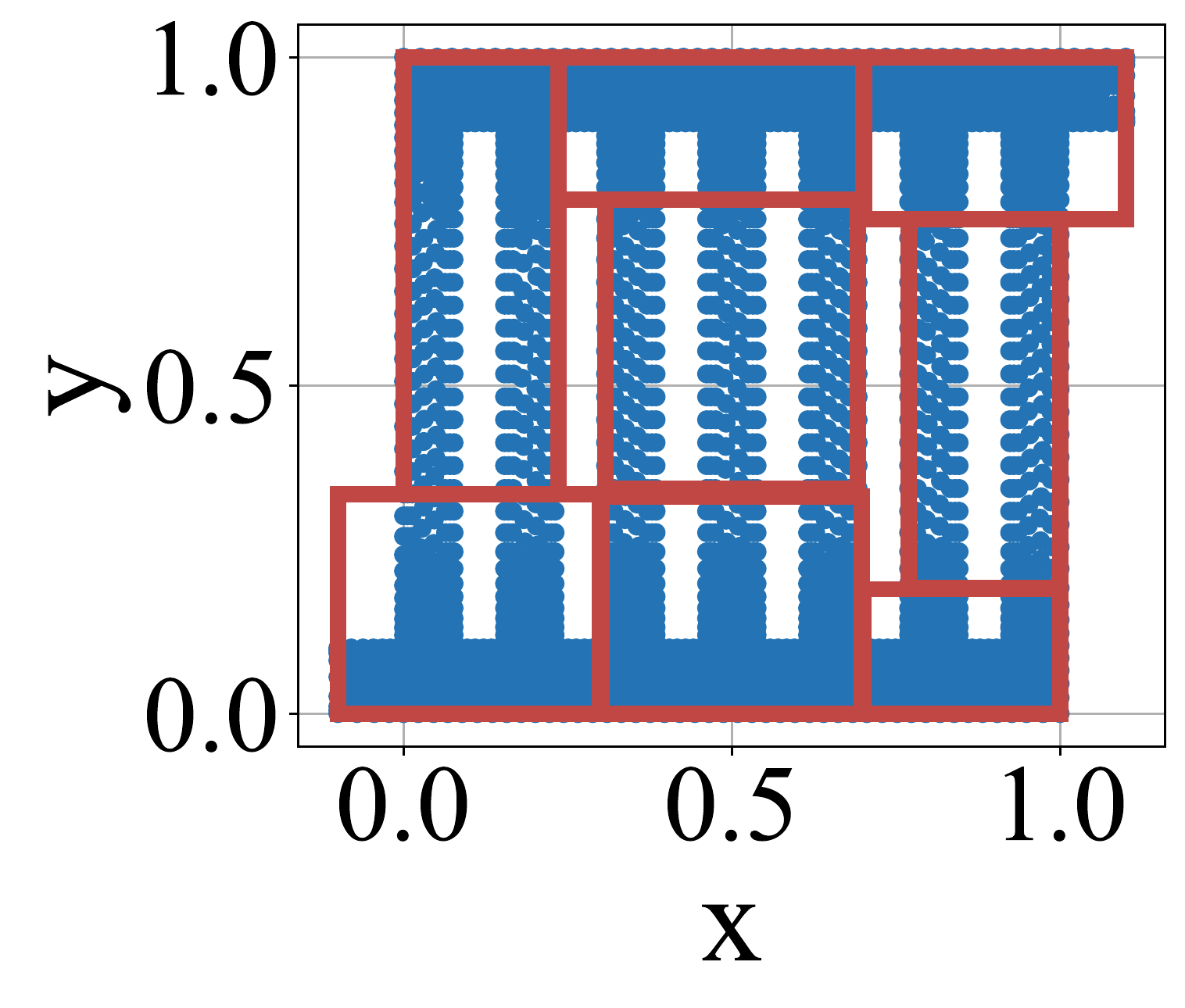}
         \caption{The geometry, point cloud, and subdomains}
         \label{fig_exp2_geom}
     \end{subfigure}
     \hfill
     \begin{subfigure}[b]{0.49\textwidth}
         \centering
         \includegraphics[height=0.29\textwidth]{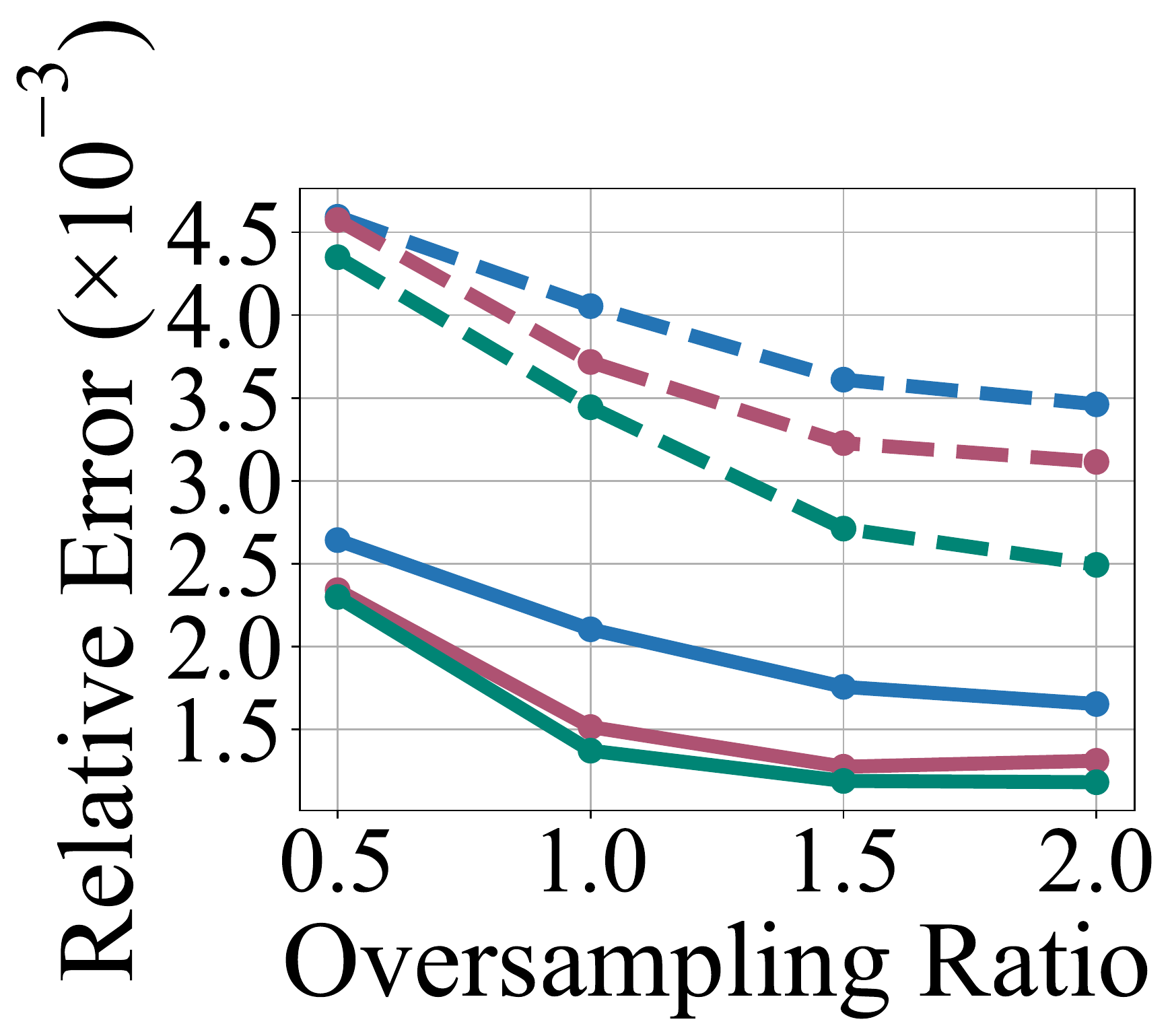}
         \hfill
         \includegraphics[height=0.27\textwidth]{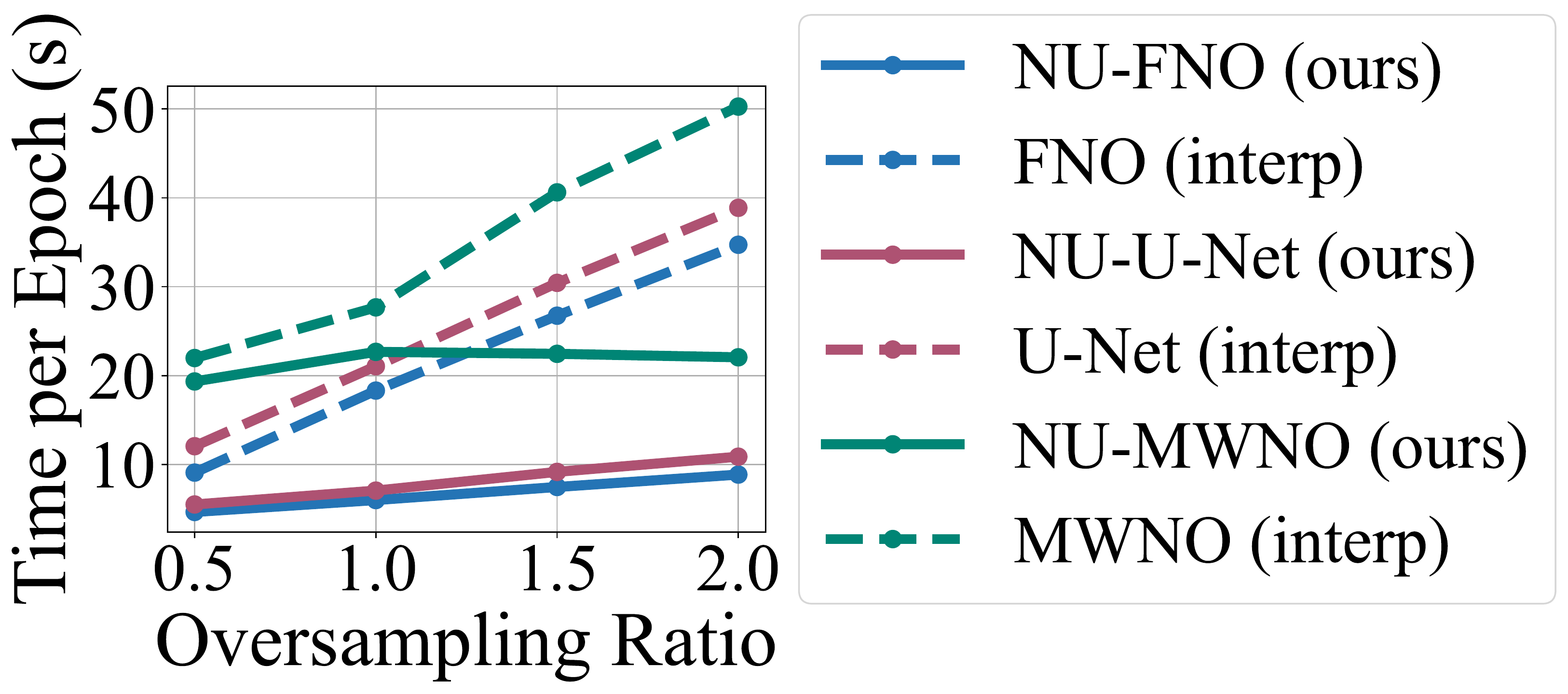}
         \caption{$L^2$ relative error (average) and training time per epoch}
         \label{fig_exp2_ratio}
     \end{subfigure}
\caption{Visualization of (2+1)D channel flow. \textbf{(a)} Gray solid rectangles (left) indicate walls between channels while red hollow rectangles (right) indicate bounding boxes of point clouds on each subdomain. \textbf{(b)} Each value of the oversampling ratio is run only once, and the random seed is $2023$. The ``interp" is an abbreviation of ``global interpolation".}
\label{fig_exp2}
\end{figure*}

\begin{table*}[t]
\centering
\caption{Experimental results of (2+1)D channel flow ($\mathrm{oversampling \ ratio}=1.5$).}
\renewcommand{\arraystretch}{1.2}
\begin{small}
\begin{tabular}{l|l|llll}
\hline
\multirow{2}{*}{\textbf{Method}} &
  \multirow{2}{*}{\textbf{\begin{tabular}[c]{@{}l@{}}Training Time\\ per Epoch ($\mathrm{s}$)\end{tabular}}} &
  \multicolumn{4}{l}{\textbf{Testing $L^2$ Relative Error ($\times 10^{-3}$)}} \\ \cline{3-6} 
 &
   &
  \multicolumn{1}{l|}{\textbf{$u$}} &
  \multicolumn{1}{l|}{\textbf{$v$}} &
  \multicolumn{1}{l|}{\textbf{$p$}} &
  \textbf{Average} \\ \hline
NU-FNO (\textbf{ours}) &
  $\bm{7.2}$ &
  \multicolumn{1}{l|}{$3.04 \pm 0.13$} &
  \multicolumn{1}{l|}{$1.34 \pm 0.04$} &
  \multicolumn{1}{l|}{$0.83 \pm 0.03$} &
  $1.74 \pm 0.05$ \\
FNO (global interpolation) &
  $26.9$ &
  \multicolumn{1}{l|}{$5.92 \pm 0.25$} &
  \multicolumn{1}{l|}{$3.52 \pm 0.09$} &
  \multicolumn{1}{l|}{$1.39 \pm 0.24$} &
  $3.61 \pm 0.19$ \\
NU-U-Net (\textbf{ours}) &
  $8.6$ &
  \multicolumn{1}{l|}{$\bm{2.18} \pm \bm{0.04}$} &
  \multicolumn{1}{l|}{$1.11 \pm 0.04$} &
  \multicolumn{1}{l|}{$0.53 \pm 0.03$} &
  $1.27 \pm 0.03$ \\
U-Net (global interpolation) &
  $29.5$ &
  \multicolumn{1}{l|}{$5.50 \pm 0.25$} &
  \multicolumn{1}{l|}{$3.41 \pm 0.17$} &
  \multicolumn{1}{l|}{$0.97 \pm 0.03$} &
  $3.29 \pm 0.15$ \\
  NU-MWNO (\textbf{ours}) &
  $22.9$ &
  \multicolumn{1}{l|}{$2.26 \pm 0.08$} &
  \multicolumn{1}{l|}{$\bm{1.05} \pm \bm{0.07}$} &
  \multicolumn{1}{l|}{$\bm{0.35} \pm \bm{0.05}$} &
  $\bm{1.22} \pm \bm{0.06}$ \\ 
  MWNO (global interpolation) &
  $40.6$ &
  \multicolumn{1}{l|}{$4.57 \pm 0.99$} &
  \multicolumn{1}{l|}{$2.76 \pm 0.81$} &
  \multicolumn{1}{l|}{$0.80 \pm 0.14$} &
  $2.71 \pm 0.63$ \\ \hline
Geo-FNO (learned) &
  $114.9$ &
  \multicolumn{1}{l|}{$2.78 \pm 1.31$} &
  \multicolumn{1}{l|}{$2.75 \pm 1.00$} &
  \multicolumn{1}{l|}{$0.93 \pm 0.41$} &
  $2.15 \pm 0.83$ \\
GraphNO &
  $193.6$ &
  \multicolumn{1}{l|}{$44.79 \pm 0.00$} &
  \multicolumn{1}{l|}{$44.17 \pm 0.00$} &
  \multicolumn{1}{l|}{$22.08 \pm 0.01$} &
  $37.02 \pm 0.00$ \\
DeepONet &
  $235.8$ &
  \multicolumn{1}{l|}{$96.86 \pm 13.45$} &
  \multicolumn{1}{l|}{$234.75 \pm 51.51$} &
  \multicolumn{1}{l|}{$27.98 \pm 6.82$} &
  $119.86 \pm 15.82$ \\ \hline
\end{tabular}
\end{small}
\label{tab_exp2}
\end{table*}

\section{Experiments}\label{sec_exp}
We consider a benchmark problem of 2D elasticity in \citet{li2022fourier} (Section~\ref{sec_exp1}), a (2+1)D problem of channel flow (Navier-Stokes equations, Section~\ref{sec_exp2}), and a challenging 3D multi-physics problem of heatsink (Section~\ref{sec_exp3}) to showcase the efficiency as well as the accuracy of our method against mesh-less and mesh-based baselines. We refer to Appendix~\ref{app_exp} for experimental details.

\begin{table}[!b]
\centering
\caption{Results of an ablation study on the number of subdomains ($\mathrm{oversampling \ ratio}=1.5$). We run each choice once with the random seed $2023$.}
\renewcommand{\arraystretch}{1.2}
\begin{small}
\begin{tabular}{l|l|llll}
\hline
 & \multirow{2}{*}{\textbf{Method}} & \multicolumn{4}{c}{\textbf{Number of Subdomains}}                                                                  \\ \cline{3-6} 
 &                                  & \multicolumn{1}{l|}{\textbf{4}} & \multicolumn{1}{l|}{\textbf{8}} & \multicolumn{1}{l|}{\textbf{12}} & \textbf{16} \\ \hline
\multirow{3}{*}{\textbf{\begin{tabular}[c]{@{}l@{}}$L^2$ Rel. Err. \\ ($\times 10^{-3}$) \\ Average\end{tabular}}} &
  NU-FNO &
  \multicolumn{1}{l|}{$1.92$} &
  \multicolumn{1}{l|}{$1.71$} &
  \multicolumn{1}{l|}{$1.31$} &
  $\bm{0.85}$ \\
 & NU-U-Net                         & \multicolumn{1}{l|}{$1.33$}     & \multicolumn{1}{l|}{$1.27$}     & \multicolumn{1}{l|}{$1.21$}      & $\bm{0.88}$ \\
 & NU-MWNO                          & \multicolumn{1}{l|}{$1.23$}     & \multicolumn{1}{l|}{$1.19$}     & \multicolumn{1}{l|}{$1.06$}      & $\bm{0.72}$ \\ \hline
\multirow{3}{*}{\textbf{\begin{tabular}[c]{@{}l@{}}Training \\ Time per \\ Epoch ($\mathrm{s}$)\end{tabular}}} &
  NU-FNO &
  \multicolumn{1}{l|}{$9.5$} &
  \multicolumn{1}{l|}{$7.2$} &
  \multicolumn{1}{l|}{$6.7$} &
  $\bm{5.7}$ \\
 & NU-U-Net                         & \multicolumn{1}{l|}{$11.2$}     & \multicolumn{1}{l|}{$8.6$}      & \multicolumn{1}{l|}{$8.0$}       & $\bm{6.9}$  \\
 & NU-MWNO                          & \multicolumn{1}{l|}{$27.6$}     & \multicolumn{1}{l|}{$23.9$}     & \multicolumn{1}{l|}{$23.1$}      & $\bm{22.7}$ \\ \hline
\end{tabular}
\end{small}
\label{tab_exp2_abla}
\end{table}

\subsection{Experiment Setup}
\paragraph{Evaluation.} 
Consistent with previous work \cite{li2020fourier,li2022fourier,lu2022comprehensive}, we use $L^2$ relative error (abbreviated as $L^2\mathrm{RE}$) to evaluate the performance of models, which can be described as:
\begin{equation}
    L^2\mathrm{RE} := \frac{1}{N}\sum_{j=1}^N \sqrt{\frac{\sum_{i=1}^M \left(u_j(y^{(i)}) - \hat{u}_j(y^{(i)})\right)^2}{\sum_{i=1}^M u_j(y^{(i)})^2}},
\end{equation}
where $y^{(i)}$ is the coordinate and $u_j, \hat{u}_j$ are the ground truth and prediction, respectively. For each experiment, we report the mean and $95\%$ confidence intervals of $L^2$ relative error after $5$ independent runs.

\paragraph{Mesh-based Benchmarks:}
\textbf{U-Net:} a famous model for approximating image-to-image mappings with spatial convolutions and deconvolutions \cite{ronneberger2015u}. 
\textbf{FNO:} a powerful neural operator with FFT-based spectral convolutions \cite{li2020fourier}.
\textbf{MWNO:} a novel neural operator with multiwavelet-based spectral convolutions \cite{gupta2021multiwavelet}.
\textbf{NU-NO:} the newly proposed framework for facilitating mesh-based neural operators with non-uniform inputs. In the following, for example, ``NU-FNO" indicates that our framework adopts an FNO as the underlying model.


\paragraph{Mesh-less Benchmarks:}
\textbf{DeepONet:} a neural operator that takes the coordinate and the parameter function as input \cite{lu2021learning}. \textbf{GraphNO:} a neural operator where both input and output are represented as graphs \cite{li2020neural}.
\textbf{Geo-FNO:} an improved version of FNO that learns to deform the input irregular mesh into a uniform mesh \cite{li2022fourier}.
\textbf{PointNet:} a well-known deep learning model for large 3D point clouds \cite{qi2017pointnet}.

\begin{table*}[t]
\centering
\caption{Experimental results of 3D heatsink.}
\renewcommand{\arraystretch}{1.2}
\begin{small}
\begin{tabular}{l|r|l|l|l}
\hline
\textbf{Method} &
  \multicolumn{1}{l|}{\textbf{Parameters}} &
  \textbf{\begin{tabular}[c]{@{}l@{}}Training Time \\ per Epoch\end{tabular}} &
  \textbf{\begin{tabular}[c]{@{}l@{}}Training Time \\ per Run\end{tabular}} &
  \textbf{\begin{tabular}[c]{@{}l@{}}Testing $L^2$ Relative \\ Error ($\times 10^{-2}$)\end{tabular}} \\ \hline
NU-FNO (\textbf{ours})                & $3,285,856$  & $\bm{4.7} \,\mathrm{s}$   & $\bm{38.9} \,\mathrm{min}$     & $\bm{5.09} \pm \bm{0.48}$    \\
FNO (global interpolation)   & $3,283,741$  & \qty{6.2}{\s}  & \qty{51.7}{\min}     & $7.68 \pm 0.25$    \\
NU-U-Net (\textbf{ours})              & $22,551,792$ & \qty{5.2}{\s}   & \qty{43.4}{\min}     & $6.31 \pm 0.31$    \\
U-Net (global interpolation) & $22,549,857$ & \qty{7.4}{\s}  & \qty{1.0}{\hour}     & $8.27 \pm 0.14$    \\
NU-MWNO (\textbf{ours})               & $6,148,715$  & \qty{18.3}{\s}   & \qty{2.5}{\hour}     & $5.36 \pm 0.24$    \\
MWNO (global interpolation)  & $6,145,565$  & \qty{21.0}{\s}   & \qty{2.9}{\hour}     & $7.38 \pm 0.12$    \\ \hline
PointNet                     & $1,677,221$  & \qty{45.3}{\s} & \qty{6.3}{\hour} & $56.42 \pm 27.60$ \\ \hline
\end{tabular}
\end{small}
\label{tab_exp3}
\end{table*}

\subsection{2D Elasticity}\label{sec_exp1}
In our first experiment, we revisit the benchmark of 2D hyper-elastic material in \citet{li2022fourier}. The governing equation is given by:
\begin{equation}
    \rho^s \frac{\partial^2 \bm{u}}{\partial t^2} + \nabla\cdot \bm{\sigma} = 0,
\end{equation}
where $\bm{\sigma}$ denotes the stress tensor. The geometry $\Omega$ is a unit square with a void of arbitrary shape at the center as shown in Figure~\ref{fig_exp1}. Our objective is to learn an operator capable of mapping the shape of the void to the values of $\bm{\sigma}$ within a point cloud of approximately $1000$ points. For more comprehensive settings, we direct the reader to the work of \citet{li2022fourier}.

We reproduce the baselines in \citet{li2022fourier} with $5$ runs and employ the proposed framework in FNO (denoted as ``NU-FNO'') to benchmark against ``Geo-FNO''. Here, the geometry $\Omega$ varies within a limited range for different inputs. So, we stack all point clouds to generate a unified domain decomposition of $6$ subdomains for all ($1000$ training and $200$ testing) data, which takes $50$ seconds.

\paragraph{Results.} The results are reported in Table~\ref{tab_exp1}. Because of reduction in interpolation error, our non-uniform framework has paralleled the accuracy and efficiency. Specifically, as shown in Figure~\ref{fig_exp1} (one of testing data, $1$ run, random seed is $2023$), FNO has a large prediction error at the edge of the void, because $\bm{\sigma}$ changes sharply here, and rough grids cannot capture such fine features. In contrast, our method uses grids of different densities in different subdomains, which can better capture features of various scales and achieve high accuracy even in places with a high variation of $\bm{\sigma}$.

\begin{figure*}[t]
     \centering
     \begin{subfigure}[b]{0.33\textwidth}
         \centering
         \includegraphics[width=.8\textwidth]{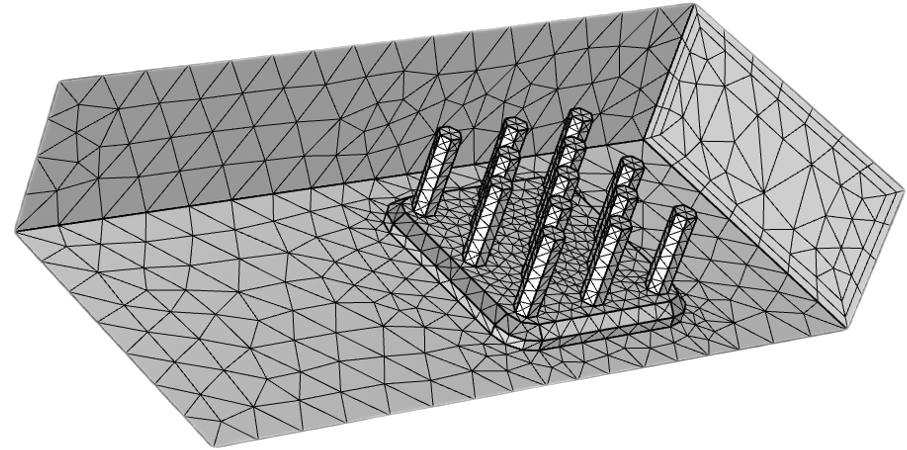}
         \vspace{0.25in}
         \caption{Complex geometry of 3D heat sink}
         \label{fig_exp3_1}
     \end{subfigure}
     \hfill
     \begin{subfigure}[b]{0.33\textwidth}
         \centering
         \includegraphics[width=.8\textwidth]{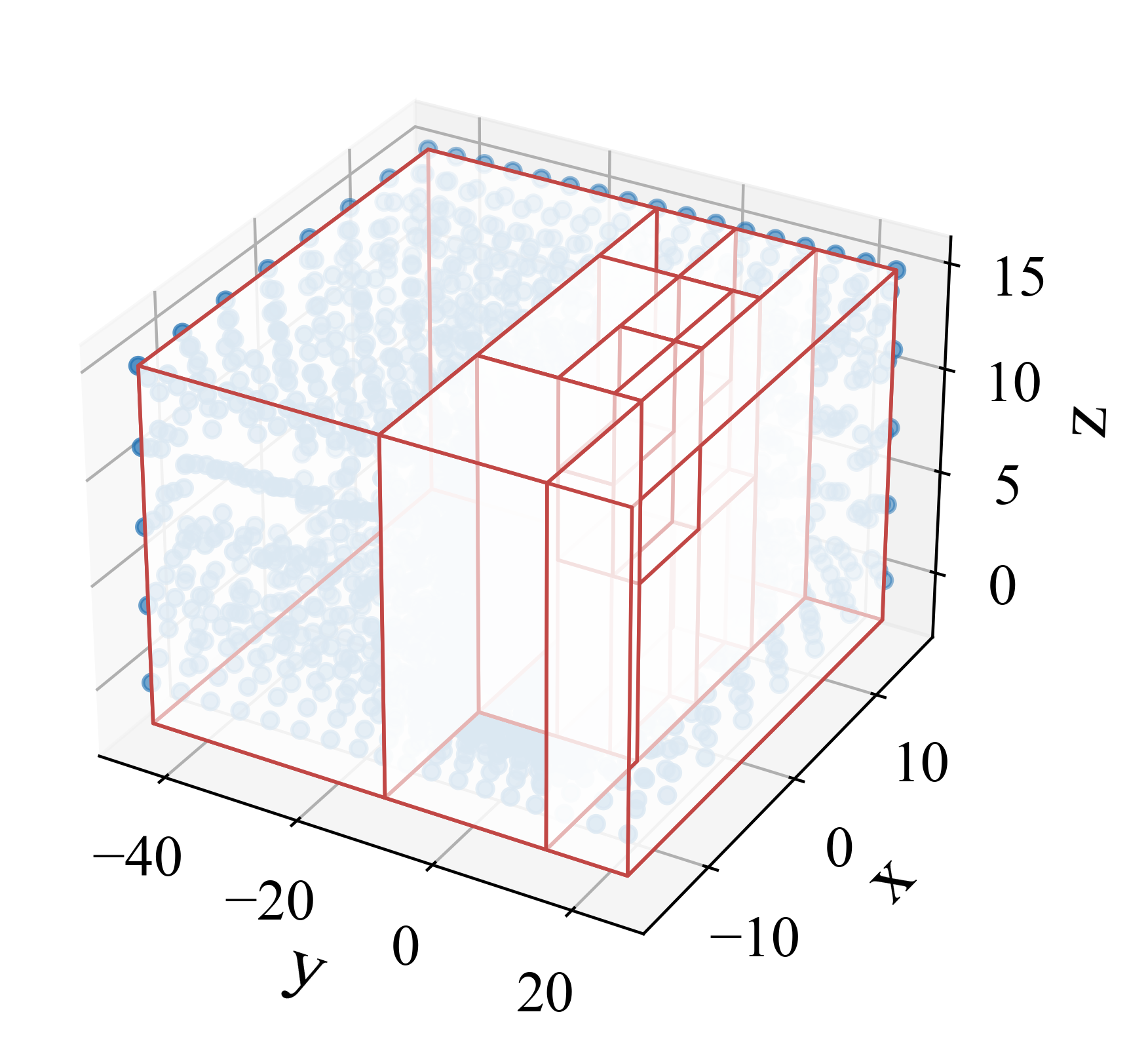}
         \caption{Point cloud and subdomains (3D)}
         \label{fig_exp3_2}
     \end{subfigure}
     \hfill
     \begin{subfigure}[b]{0.33\textwidth}
         \centering
         \includegraphics[width=.8\textwidth]{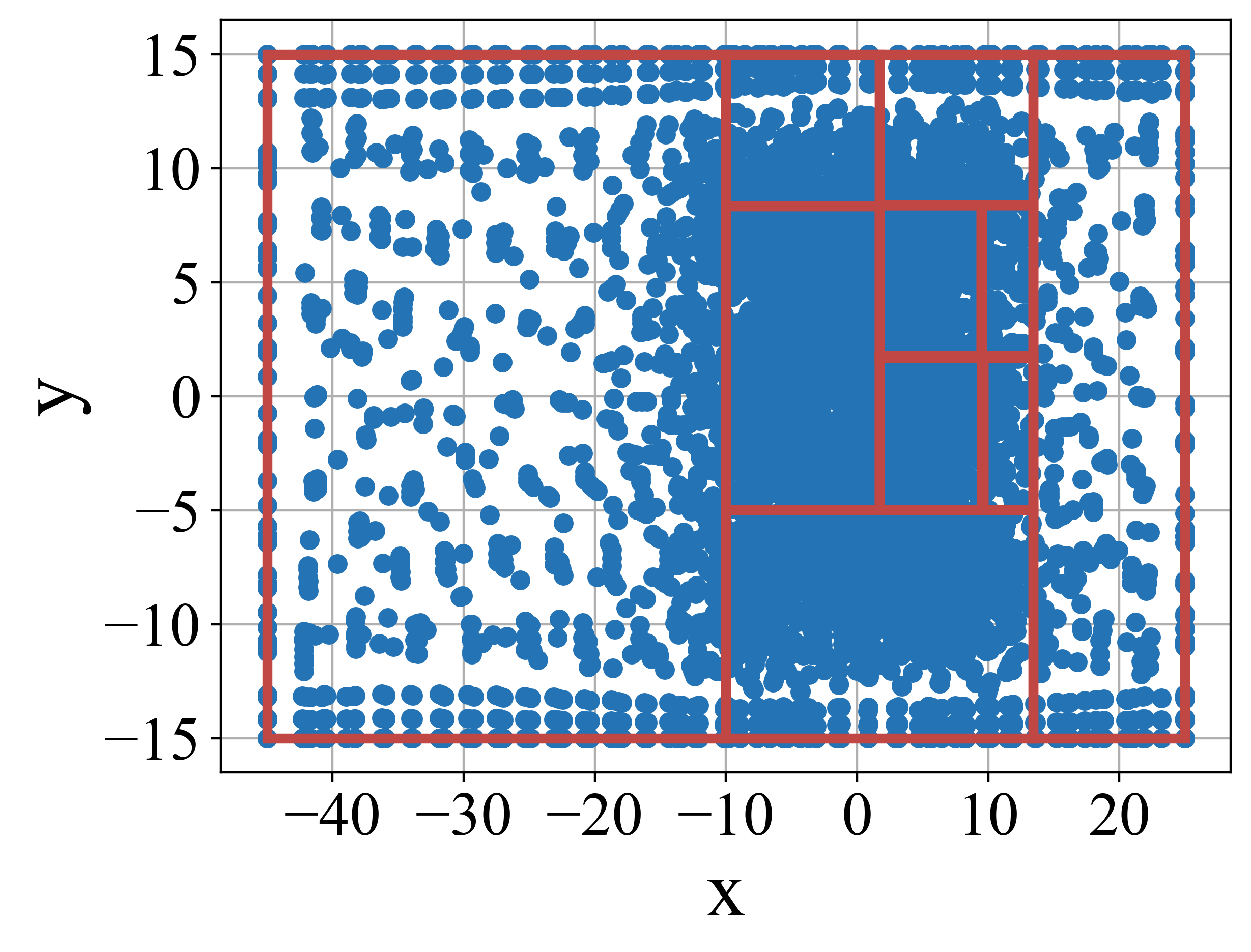}
         \caption{Point cloud and subdomains (2D)}
         \label{fig_exp3_3}
     \end{subfigure}
\caption{Visualization of 3D heatsink. \textbf{(a)} The model of heatsink comes from real-world application. \textbf{(b)} Red lines indicate the borders of subdomains. \textbf{(c)} 2D projection (to the XY plane) of the 3D point cloud and subdomains.}
\label{fig_exp3}
\end{figure*}

\subsection{(2+1)D Channel Flow}\label{sec_exp2}
We consider the 2D time-dependent incompressible flow in a row of channels adapted from \citet{aparicio2020cfd}, which is governed by the notoriously hard-to-solve (incompressible) Navier-Stokes equations:
\begin{equation}\label{eq_exp2}
\begin{aligned}
    \frac{\partial \bm{u}}{\partial t} + \bm{u} \cdot \nabla \bm{u} &= -\nabla p  + \nu\nabla^2\bm{u},\\
    \nabla \boldsymbol \cdot \bm{u} &= 0,\\
\end{aligned}
\end{equation}
where $x\in\Omega,t\in[0,0.3)$ are space and time coordinates, respectively, $\bm{u}=(u(x,t),v(x,t))$ is the velocity field, $p=p(x,t)$ is the pressure, and $\nu$ is the viscosity. As displayed in Figure~\ref{fig_exp2_geom}, the geometry $\Omega\subset\mathbb{R}^2$ consists of an inlet, an outlet, and a row of channels. For this case, the target operator is defined to be a mapping: $(u, v, p)|_{x\in\Omega, t\in[0,0.15)}\mapsto (u, v, p)|_{x\in\Omega, t\in[0.15,0.3)}$. 

We produce a dataset consisting of $1000$ training instances and $200$ test instances. The temporal interval $[0, 0.3)$ is discretized into $30$ distinct steps, whereas the spatial domain is represented by a point cloud comprising $3809$ points. Consequently, the total count of points within the point cloud amounts to $30 \times 3809 = 114,270$. Pertaining to mesh-free baselines, the inputs comprise the $(u, v, p)$ values situated at the locations defined by the point cloud during the time interval $t\in [0,0.15)$. For mesh-based baselines, the input data is constituted by an interpolated mesh of size $\mathrm{oversampling \ ratio}\times 3809$ during $t\in [0,0.15)$. In contrast, the input for our proposed method consists of meshes spanning $8$ subdomains (refer to Figure~\ref{fig_exp2_geom}), with a total size of $\mathrm{oversampling \ ratio}\times 3809$ during the same time interval. Given that $\Omega$ remains invariant w.r.t. the input, we only necessitate a single domain decomposition, requiring a mere $0.5$ seconds.

\paragraph{Results.} The testing results are given in Table~\ref{tab_exp2}. Compared with mesh-based and mesh-free baselines, our method has achieved both state-of-the-art performance and the highest speed. This is because our method uses a grid adaptive to the point cloud distribution density in each subdomain, which reduces the interpolation error and encourages the neural operators to extract multi-scale features. Since the grid size of the subdomain is much smaller than that from global interpolation, the size of hidden embeddings is also reduced (the number of hidden-layer channels remains the same), allowing for a more compact representation and higher training speed.

\paragraph{Mesh Size.} We re-run mesh-based methods under different mesh sizes (or equivalently, oversampling ratios). As shown in Figure~\ref{fig_exp2_ratio}, even with $\mathrm{oversampling \ ratio}=0.5$, our method significantly outperforms global interpolation with $\mathrm{oversampling \ ratio}=2.0$. In addition, we also find that the error decline of global interpolation slows down as the grid gets denser, showing a trend that it is difficult to outperform our method. This gap may be caused by the stronger ability of our method to extract multi-scale features. Besides, due to the compact representation, the time cost of our method grows far slower with mesh size than with global interpolation.

\paragraph{Ablation Study.} We evaluated our methods (with various underlying models) under a range of choices for the number of subdomains. From the results shown in Table~\ref{tab_exp2_abla}, we unexpectedly discover both an increase in accuracy and speed as the number of subdomains increases, which seems like the ``blessing of subdomains''. We attribute this to the following reasons: firstly, an increase in the number of subdomains results in lower interpolation error and yields finer features, thereby enhancing performance. Secondly, the most time-intensive step in our framework is the third step, which involves the hidden layers of the neural operator (as detailed in Section~\ref{sec_nuno}). As we have maintained a fixed total mesh size, an increase in the number of subdomains leads to a decrease in the mesh size for each individual subdomain (i.e., $s\downarrow$). Given a fixed hidden width $n_h$, this results in a smaller size for the hidden embeddings (i.e., $n_h\times s\downarrow$), thereby accelerating the forward pass. By expanding the number of subdomains, we reduce redundancy in the hidden embeddings, enabling a more compact representation. 


\subsection{3D Heatsink}\label{sec_exp3}
As a fresh attempt in the field of the neural operator, we consider a 3D heatsink from the COMSOL application gallery \cite{comsol} in the last experiment. The governing equations consist of not only Navier-Stokes equations for fluid dynamics modeling, but also a series of equations for heat transfer modeling: conduction, convection, radiation, and the heat equation. For brevity, we refer to \citet{haertel2015topology} for details of governing equations. The geometry $\Omega\subset \mathbb{R}^3$ is a rectangular enclosure with a heatsink inside as shown in Figure~\ref{fig_exp3_1}. The physical quantities contain the velocity field $\bm{u} = (u(x), v(x), w(x)), x\in \Omega$, the pressure field $p = p(x)$, and the temperature field $T=T(x)$. Our interested operator maps the velocity field $\bm{u}$ to the resulting temperature $T$.

We generate $900$ training and $100$ testing data. The 3D point cloud contains $19,517$ points. Since previously considered mesh-free neural operators scale badly to 3D point clouds, we consider another mesh-free model, PointNet, which is designed for analyzing large 3D point clouds. For mesh-based baselines, the input is the interpolated mesh of the size aligned with the point cloud size and of the same aspect ratio as the geometry $\Omega$. For our method, as shown in Figure~\ref{fig_exp3_2} and \ref{fig_exp3_3}, the geometry is decomposed into $16$ subdomains, taking $8$ seconds. Subsequently, the point cloud is interpolated into uniform grids over subdomains, which are used for training our model.

\paragraph{Results.}
The challenge of this problem is that higher spatial degrees of freedom will significantly slow down the mesh-free neural operators, and point clouds that are non-uniformly distributed in 3D space are more difficult to be accurately interpolated to a uniform grid. Through domain decomposition, our method utilizes a set of grids with varied sizes to approximate the distribution of the point cloud, which is much more efficient than global interpolation. This is why our method is so accurate and high-speed in this 3D problem, as reported in Table~\ref{tab_exp3}. Moreover, we note that the parameter size of our method is only slightly larger than the original baseline. This is because when applying our framework, only additional projections $P$ and $Q$ are needed, whose parameter size is negligible. 


\section{Conclusion and Future Work}
In this paper, we propose a framework named non-uniform neural operator (NUNO) to handle the challenge of non-uniform data in real-world applications. Through K-D tree-based domain decomposition and subsequent interpolation, we convert non-uniform data into uniform grids, allowing for fast mesh-based techniques such as the FFT while persevering a low interpolation error. In experiments, we benchmark on 2D elasticity, (2+1)D channel flow, and 3D heatsink. Results show that our framework has achieved both state-of-the-art accuracy and promising efficiency. Even at the coarsest resolution, our method still outperforms mesh-based baselines which are at the finest resolution.

\textbf{3D Complex Geometry.} Our work is a first work to investigate PDE problems of 3D complex geometries. However, there is still a lot of work to be done to identify and address potential challenges. For example, we may learn from computer graphics and design special architectures for large point clouds.  \textbf{Uniform Grids of Varied Sizes.} In this work, we adopt the simplest technique, i.e., resizing, to deal with uniform grids of different sizes. It is interesting to explore other approaches such as SPP-Net \cite{he2015spatial}. \textbf{Variable-Structure Problems.} One of the limitations of our work is that it is only applicable to the problems where the geometry does not change with the input $a$ or does change within a limited range. In the other cases, we must generate different domain divisions for different inputs, to which the projections $P$ and $Q$ cannot generalize. As a future direction, we will try to combine our framework with GNNs and others, expediting applications in relevant fields such as structural optimization \cite{christensen2008introduction}.

\section*{Acknowledgements}
This work was supported by the National Key Research and Development Program of China (2020AAA0106000, 2020AAA0104304), NSFC Projects (Nos. 62061136001, 62076145, 62076147, U19B2034, U1811461, U19A2081, 61972224), BNRist (BNR2022RC01006), Tsinghua Institute for Guo Qiang, and the High Performance Computing Center, Tsinghua University. J.Z is also supported by the New Cornerstone Science Foundation through the XPLORER PRIZE.


\bibliography{paper}
\bibliographystyle{icml2023}

\newpage
\appendix
\onecolumn

\section*{Limitations and Broader Impacts}
In this work, we mainly discuss the challenges of non-uniform data for neural operators. And we develop a general framework named non-uniform neural operator to tackle these challenges. Since this work is basic research, we think there are no relevant ethical issues.

\section{Appendix}


\subsection{Complexity Analysis of K-D Tree Algorithm}\label{app_complexity}
In this subsection, we will analyze the time and space complexity of Algorithm~\ref{alg_1}. For ease of reference, we have replicated Algorithm~\ref{alg_1} into Algorithm~\ref{alg_2}.

\paragraph{Assumptions.} Let $m$ be the number of points in the initial point cloud, i.e., $m:=|D^{(0)}|$. Before the analysis, we make the following assumptions:
\begin{itemize}
    \item To ensure that the problem is well defined, we assume that the number of sub-point clouds cannot exceed that of points in the initial point cloud, i.e., $n\le m$.
    \item To ensure that the statistics in Eq.~\eqref{eq_uniform} are reasonable, we assume that $N_1\times\cdots\times N_d \le |D|$. Since $|D|\leq |D^{(0)}|$ (the number of points in the sub-point cloud will never exceed the initial point cloud), we have that $N_1\times\cdots\times N_d \le m$.
    \item In Line~8, we will not necessarily investigate all possible candidates of $b$ (i.e., to enumerate all locations of the points in $D$) but consider $N_{\mathrm{max}}$ discrete equidistant candidates. We assume $N_{\mathrm{max}}$ to be a typical small constant, e.g., $N_{\mathrm{max}}=5$, which is consistent with our implementation in experiments.
\end{itemize}

\paragraph{Time Complexity.} Firstly, we discuss the cost of evaluating Eq.~\eqref{eq_uniform}. During the execution of the algorithm, sub-point clouds that ever existed (including intermediates that were later decomposed) form a full binary tree with $2n - 1$ nodes. For each node, Eq.~\eqref{eq_uniform} is only evaluated once, whose time complexity is $\mathcal{O}(|D| + d)$, noticing that $N_1\times\cdots\times N_d \le |D|$. Since $|D|\le m$, the total cost of Eq.~\eqref{eq_uniform} comes as $\mathcal{O}(nm + nd)$. Secondly, we analyze the cost of the loops. The number of sub-point clouds will increase by one each time the loop runs, so the number of loops is exactly $n-1$. In each loop, Line~5, 6, 10 cost no more than $\mathcal{O}(n)$ (the cost of obtaining $\infdiv{P}{Q}{D}$ by Eq.~\eqref{eq_uniform} has been calculated in the previous step); Line~7 cost $\mathcal{O}(d)$; Line~9 is nothing but a duplication, which is $\mathcal{O}(|D|)$. As for Line~8, we need to sort the point cloud along the $k$-th axis before evaluating Eq.~\eqref{eq_uniform_after} for $N_{\mathrm{max}}$ candidates, which cost no more than $\mathcal{O}(|D|\log |D| + d)$. To sum up, since $|D|\le m$ and $n\le m$, the time complexity of loops is $\mathcal{O}(nm\log m + nd)$ which is also the total time complexity.

\paragraph{Space Complexity} In the lifetime of the algorithm, there is only a reorganization of point clouds, and no new point clouds are generated. So $|\bigcup_{D\in\mathcal{S}} D|=|D^{(0)}|$ holds all the time and the space complexity is simply $\mathcal{O}(md)$, where $d$ comes from the length of the spatial coordinates of each point.

\begin{algorithm}[!b]
   \caption{Domain Decomposition via a K-D Tree (replica)}
   \label{alg_2}
\begin{algorithmic}[1]
   \STATE {\bfseries Input:} initial point cloud $D^{(0)}$ and the number of sub-point clouds $n$
   \STATE {\bfseries Output:} a sef of sub-point clouds $\mathcal{S}$
   \STATE {\bfseries Initialize:} $\mathcal{S} \leftarrow \{ D^{(0)} \}$
   \REPEAT
   \STATE Choose $D^* = \argmax_{D\in \mathcal{S}} (|D| \cdot \infdiv{P}{Q}{D})$
   \STATE $\mathcal{S}\leftarrow\mathcal{S}-\{ D^* \}$
   \STATE Select the dimension $k, 1\le k \le d$, where the bounding box of $D^*$ has the largest scale
   \STATE Determine the hyperplane $x_k = b^*$, where $b^* = \argmax_{b\in \mathcal{B}} \mathrm{Gain}(D^*, b)$ and $\mathcal{B}$ is a set of discrete candidates for $b$
   \STATE Partition $D^*$ with $x_k = b^*$
   \STATE $\mathcal{S}\leftarrow\mathcal{S}\cup\{ D^*_{x_k>b^*}, D^*_{x_k\le b^*} \}$
   \UNTIL{$|\mathcal{S}|=n$}
\end{algorithmic}
\end{algorithm}

\subsection{Experimental Details}\label{app_exp}
In this part, We provide supplementary descriptions of the experimental environment, governing equations, hyperparameters, and implementation details.

\paragraph{Experimental environment.}
Our codes are written in Python and the deep learning part is based on PyTorch library. We refer to \texttt{README} in our code repository for the details of dependencies and instructions. All the codes are run on Ubuntu 18.04 LTS Intel(R) Xeon(R) with $32$ Intel(R) Xeon(R) CPU E5-2620 v4 @ 2.10GHz CPUs. All the models are trained on $1$ NVIDIA TITAN Xp 12GB GPU. Besides, we use COMSOL Multiphysics as the software of numerical solver for data generation.

\subsubsection{2D Elasticity}\label{app_exp2}
\paragraph{Data Genearation.} The dataset directly comes from \citet{li2022fourier}. The settings and parameters of the PDEs are identical to the original experiment in \citet{li2022fourier}.

\paragraph{Implementation of Our Method.} The input of the operator is the void shape (see Figure~\ref{fig_exp1} for an example) which can be naturally represented by the point cloud. So, we use the \texttt{KernelDensity} (of the bandwidth $0.05$) from the scikit-learn to approximate the distribution of the point cloud. Then, we evaluate the likelihood at the $32\times 32$ uniform grid, which serves as the network input. The network outputs the prediction for $\bm{\sigma}$ values at the subdomain grids, which are interpolated back to the point cloud for evaluation.

\paragraph{Hyperparameters.} The hyperparameters of the baselines remain the same as in the original paper, except for three modifications:
\begin{itemize}
    \item To accelerate the training process, the batch size for the DeepONet baseline is increased from $16$ to $16384$.
    \item In the original implementation, for the mesh-based neural operators, the error is measured by the deviation between the output of model and the mesh interpolated from the ground-truth labels at the point cloud. We think this may be inappropriate since our target is to predict the point cloud rather than the interpolated mesh. Therefore, we have corrected the calculation of the error as the deviation between the interpolation result of the model output and the ground-truth labels at the point cloud.
    \item We changed the original single experiment to parallel five experiments.
\end{itemize}
For our method, the optimizer is \texttt{Adam} with learning rate $0.001$ and weight decay $10^{-4}$. The training period consists of $501$ training epochs with batch size $20$. The learning rate scheduler is \texttt{StepLR} with step size $400$ and gamma $0.1$. In addition, NU-FNO uses $4$ Fourier layers with modes $12$ and width $32$ for normal FNO processing. When it is finished, we use $2$ linear layers and $2$ Fourier layers (with modes $12$ and width $32$) to generate the prediction in subdomains and interpolate the result into the point cloud via NUDFT. The dropout and normalization are used to promote performance.

\subsubsection{(2+1)D Channel Flow}
\paragraph{Data Generation.} The governing equations are given in Eq.~\eqref{eq_exp2}, where $\nu=0.02$. The geometry is shown in Figure~\ref{fig_exp2_geom}. The initial condition is $(u(x,0),v(x,0),p(x,0))=(0,0,0), x\in\Omega$. We pose $\bm{u}(x,t)=(u_0(x), 0), t\in(0,0.3)$ at the inlet, $p(x,t)=1, t\in(0,0.3)$ at the outlet, and no-slip boundary condition at other boundaries. We generate $u_0$ samples as $u_0\sim \mu$, where $\mu = \mathcal{N}(0, 7^2(-\Delta+49I)^{-2.5})$. Then, the numerical solver is used to solve the PDEs with corresponding $u_0$ parameters to generate final training and testing data. The link of the dataset is put into the \texttt{README} in our code repository.

\paragraph{Implementation of Our Method.} The input of the operator is the $(u,v,p)$ values at the point cloud from $t=0$ to $t=0.14$. We first interpolate them into uniform grids of $8$ subdomains, which are long aligned to reduce padding. Then, the grids are resized to the same size (the largest one of their original sizes) via the FFT-IFFT and then passed to the network. The network outputs the prediction for $(u,v,p)$ values at the subdomain grids from $t=0.15$ to $t=0.29$, which are interpolated back to the point cloud for evaluation.

\paragraph{Hyperparameters.} The optimizer is \texttt{Adam} with learning rate $0.001$ and weight decay $10^{-4}$. The learning rate scheduler is \texttt{ReduceLROnPlateau} with patience $\lfloor \mathrm{epochs}/20 \rfloor$ and factor $0.1$. The hyperparameters of each baseline is listed below:
\begin{itemize}
    \item \textbf{FNO (global interpolation).} It has $4$ 3D Fourier layers with modes $8$ and width $20$. It is trained for $501$ epochs with batch size $20$.
    \item \textbf{U-Net (global interpolation).} It has $4$ 3D convolution layers and $4$ 3D deconvolution layers. It is trained for $501$ epochs with batch size $20$.
    \item \textbf{MWNO (global interpolation).} It has $4$ multiwavelet layers with $\alpha=8,c=3,k=3,L=0$ and Legendre basis. It is trained for $501$ epochs with batch size $20$.
    \item \textbf{NU-FNO, NU-U-Net, NU-MWNO.} The parameters of our method are the same as corresponding baselines, except for the input and output channels which are increased to allow for taking and producing subdomain uniform grids.
    \item \textbf{Geo-FNO (learned).} It has $2$ 2D Fourier layers with modes $8$ and width $20$ for interpolation of input and output via NUDFT and $3$ 3D Fourier layers with Fourier layers with modes $8$ and width $20$. It is trained for $101$ epochs with batch size $10$.
    \item \textbf{GraphNO.} It has $4$ graph convolution layers with node width $32$ and kernel width $128$. The input graph is built from the point cloud with a Gaussian kernel with radius $0.05$ and the resulting edge features have $4$ channels. It is trained for $101$ epochs with batch size $1$.
    \item \textbf{DeepONet.} The input coordinate of DeepONet is organized as $(x_i, y_i)$ while $(u, v, p)$ values at different time steps are considered as multiple channels, rather than $(x_i, y_i, t_i)$. Besides, it has $5$ branch layers and $5$ trunk layers, both with $64$ neurons. When branch and trunk complete their calculations, their results are grouped, where the dot product is performed in each group separately to produce the final multi-channel results. It is trained for $101$ epochs with batch size $2048$. And the learning rate is $0.0001$ which is different from other baselines.
\end{itemize}

\subsubsection{3D Heatsink}
\paragraph{Data Genearation.} We refer to the COMSOL application gallery \cite{comsol} for the details of the governing equations \footnote{\url{https://www.comsol.com/model/heat-sink-8574}}. The body force term in NS equations is denoted as $\bm{f} = (f, 0, 0)(x), x\in\Omega$, where $f(x) = f_0(x, y)$ only changes in the XY direction (here we abuse the notations for coordinates: the former $x$ is a coordinate in the three-dimensional space while the latter $x$ is the value on the X axis). We generate $f_0$ samples as $f_0\sim \mu$, where $\mu = \mathcal{N}(0, 7^{1.5}(-\Delta+49I)^{-2.5})$. Then, the numerical solver is used to solve the PDEs with corresponding $f_0$ parameters to generate final training and testing data. The link of the dataset is put into the \texttt{README} in our code repository.

\paragraph{Implementation of Our Method.} The implementation details are similar to the details in Appendix~\ref{app_exp2}.

\paragraph{Hyperparameters.} The optimizer is \texttt{Adam} with learning rate $0.001$ and weight decay $10^{-4}$. The learning rate scheduler is \texttt{ReduceLROnPlateau} with patience $\lfloor \mathrm{epochs}/20 \rfloor$ and factor $0.1$. All the models are trained for $501$ epochs with batch size $20$. The hyperparameters of each baseline is listed below:
\begin{itemize}
    \item \textbf{FNO (global interpolation).} It has $4$ 3D Fourier layers with modes $8$ and width $20$.
    \item \textbf{U-Net (global interpolation).} It has $4$ 3D convolution layers and $4$ 3D deconvolution layers.
    \item \textbf{MWNO (global interpolation).} It has $4$ multiwavelet layers with $\alpha=8,c=3,k=3,L=0$ and Legendre basis.
    \item \textbf{NU-FNO, NU-U-Net, NU-MWNO.} The parameters of our method are the same as corresponding baselines, except for the input and output channels which are increased to allow for taking and producing subdomain uniform grids.
    \item \textbf{PointNet.} It has $4$ local transform layers and a PointNet feature layer.
\end{itemize}
In addition, for mesh-based baselines, the loss function consists of two terms: the first term is the deviation between the network output and the ground-truth subdomain grids while the last term is a regularization term, which is the interpolation error of between output and the ground-truth point cloud. The weight of the regularization term is $5\times 10^{-3}$.

\paragraph{Remark.} When calculating the shape of the uniform grid, we let the total size (i.e., the number of cells in the grid) be equal to the number of points in the point cloud, $19517$, and let the aspect ratio be equal to that of the geometry. However, since the size of grid must be an integer, there are some rounding errors being ignored. 


\end{document}